\documentclass[11pt,a4paper]{article}
\usepackage[hyperref]{acl2021}
\usepackage{times}
\usepackage{latexsym}
\usepackage{times}
\usepackage{graphicx}
\usepackage{float}
\usepackage{import}
\usepackage{amsmath}
\usepackage{booktabs}
\usepackage{color}
\usepackage{amssymb}
\usepackage{url}
\usepackage{multirow}
\usepackage{caption, subcaption}
\usepackage{fancyvrb}
\usepackage{makecell}

\usepackage{pgfplots}
\usepackage{pgfplotstable}
\usepackage{tikz}

\usepackage{enumitem}
\usepackage{microtype}

\aclfinalcopy 

\newcommand\modelname{\textsc{Text2Event}}

\newcommand{\red}[1]{\textcolor{red}{#1}}

\newcommand{\eventtpye}[1]{\textit{#1}}

\title{\modelname: Controllable Sequence-to-Structure Generation \\ for End-to-end Event Extraction}

\author{
  Yaojie Lu${}^{1,3}$,
  Hongyu Lin${}^{1}$,
  Jin Xu${}^{4,}$\thanks{~ Corresponding authors.},
  Xianpei Han${}^{1,2,}$\footnotemark[1],
  Jialong Tang${}^{1,3}$,
  \\
  \textbf{Annan Li}${}^{1,3}$,
  \textbf{Le Sun}${}^{1,2}$,
  \textbf{Meng Liao}${}^{4}$,
  \textbf{Shaoyi Chen}${}^{4}$
  \\
  ${}^{1}$Chinese Information Processing Laboratory ~
  ${}^{2}$State Key Laboratory of Computer Science \\
  Institute of Software, Chinese Academy of Sciences, Beijing, China\\
  ${}^{3}$University of Chinese Academy of Sciences, Beijing, China \\
  ${}^{4}$Data Quality Team, WeChat, Tencent Inc., China \\
 {\tt \{yaojie2017,jialong2019,liannan2019\}@iscas.ac.cn} \\
 {\tt \{hongyu,xianpei,sunle\}@iscas.ac.cn} \\
 {\tt \{jinxxu,maricoliao,shaoyichen\}@tencent.com }
}
\date{}

\begin{document}
\maketitle

\begin{abstract}

Event extraction is challenging due to the complex structure of event records and the semantic gap between text and event.
Traditional methods usually extract event records by decomposing the complex structure prediction task into multiple subtasks.
In this paper, we propose \modelname, a sequence-to-structure generation paradigm that can directly extract events from the text in an end-to-end manner.
Specifically, we design a sequence-to-structure network for unified event extraction, a constrained decoding algorithm for event knowledge injection during inference, and a curriculum learning algorithm for efficient model learning.
Experimental results show that, by uniformly modeling all tasks in a single model and universally predicting different labels, our method can achieve competitive performance using only record-level annotations in both supervised learning and transfer learning settings.

\end{abstract}

\section{Introduction}\label{sec:introduction}

\begin{figure}[!tpb]
    \centering
    \includegraphics[width=0.49\textwidth]{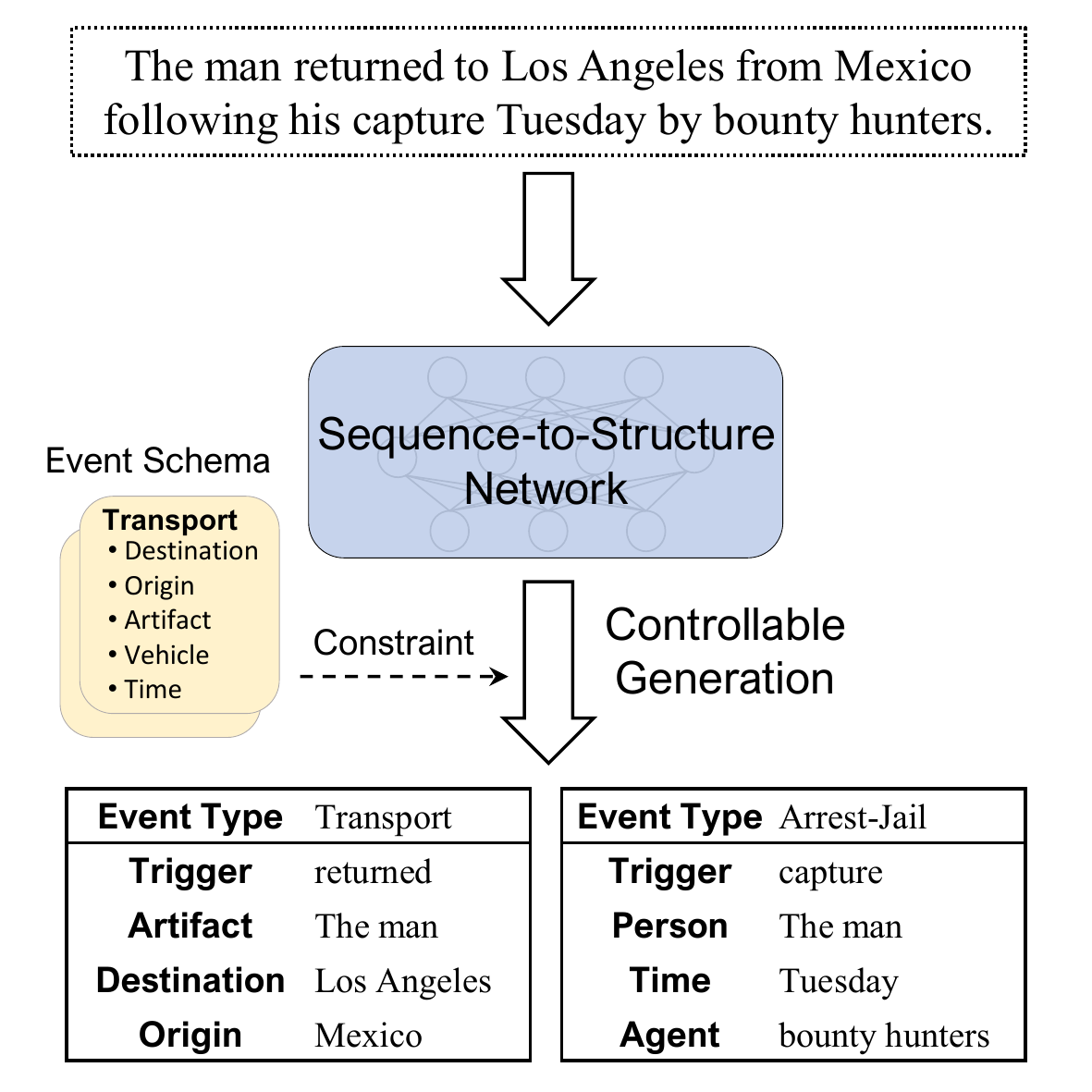}
    \caption{
        The framework of \modelname.
        Here, \modelname\, takes  raw  text as  input  and  generates a \eventtpye{Transport} event and an \eventtpye{Arrest-Jail} event.
        }
    \label{fig:motivation}
\end{figure}

Event extraction is an essential task for natural language understanding, aiming to transform the text into event records \citep{doddington-etal-2004-automatic,ahn:2006:ARTE}.
For example, in \figurename~\ref{fig:motivation}, mapping ``The man returned to Los Angeles from Mexico following his capture Tuesday by bounty hunters.'' into two event records 
\{Type: \eventtpye{Transport}, Trigger: returned, Arg1 Role: \eventtpye{Artifact}, Arg1: The man, Arg2 Role: \eventtpye{Destination}, Arg2: Los Angeles, ... \} 
and 
\{Type: \eventtpye{Arrest-Jail}, Trigger: capture, Arg1 Role: \eventtpye{Person}, Arg1: The man, Arg2 Role: \eventtpye{Agent}, Arg2: bounty hunters, ... \}.

Event extraction is challenging due to the complex structure of event records and the semantic gap between text and event.
First, an event record contains event type, trigger, and arguments, which form a table-like structure.
And different event types have different structures.
For example, in \figurename~\ref{fig:motivation}, \eventtpye{Transport} and \eventtpye{Arrest-Jail} have entirely different structures.
Second, an event can be expressed using very different utterances, such as diversified trigger words and heterogeneous syntactic structures.
For example, both ``the dismission of the man'' and ``the man departed his job'' express the same event record \{Type: \eventtpye{End-Position}, Arg1 Role: \eventtpye{PERSON}, Arg1: the man\}.

Currently, most event extraction methods employ the decomposition strategy \citep{chen-etal-2015-event,Nguyen-Nguyen:2019:AAAI2019,wadden-etal-2019-entity,Zhang:2019:GAIL,du-cardie-2020-event,li-etal-2020-event,paolini2021structured}, i.e., decomposing the prediction of complex event structures into multiple separated subtasks (mostly including entity recognition, trigger detection, argument classification), and then compose the components of different subtasks for predicting the whole event structure (e.g., pipeline modeling, joint modeling or joint inference).
The main drawbacks of these decomposition-based methods are:
(1) They need massive and fine-grained annotations for different subtasks, often resulting in the data inefficiency problem.
For example, they need different fine-grained annotations for \eventtpye{Transport} trigger detection, for \eventtpye{Person} entity recognition, for \eventtpye{Transport.Artifact} argument classification, etc.
(2) It is very challenging to design the optimal composition architecture of different subtasks manually.
For instance, the pipeline models often lead to error propagation.
And the joint models need to heuristically predefine the information sharing and decision dependence between trigger detection, argument classification, and entity recognition, often resulting in suboptimal and inflexible architectures.

In this paper, we propose a sequence-to-structure generation paradigm for event extraction -- \modelname, which can directly extract events from the text in an end-to-end manner.
Specifically, instead of decomposing event structure prediction into different subtasks and predicting labels, we uniformly model the whole event extraction process in a neural network-based sequence-to-structure architecture, and all triggers, arguments, and their labels are universally generated as natural language words.
For example, we generate a subsequence ``Attack fire'' for trigger extraction, where both ``Attack'' and ``fire'' are treated as natural language words.
Compared with previous methods, our method is more data-efficient: it can be learned using only coarse parallel text-record annotations, i.e., pairs of $\langle$sentence, event records$\rangle$, rather than fine-grained token-level annotations.
Besides, the uniform architecture makes it easy to model, learn and exploit the interactions between different underlying predictions, and the knowledge can be seamlessly shared and transferred between different components.

Furthermore, we design two algorithms for effective sequence-to-structure event extraction.
First, we propose a constrained decoding algorithm, which can guide the generation process using event schemas.
In this way, the event knowledge can be injected and exploited during inference on-the-fly.
Second, we design a curriculum learning algorithm, which starts with current pre-trained language models (PLMs), then trains them on simple event substructure generation tasks such as trigger generation and independent argument generation, finally trains the model on the full event structure generation task.

We conducted experiments\footnote{Our source codes are openly
available at \href{https://github.com/luyaojie/text2event}{https://github.com/luyaojie/text2event}} on ACE and ERE datasets, and the results verified the effectiveness of \modelname\, in both supervised learning and transfer learning settings.
In summary, the contributions are as follows:
\begin{enumerate}
    \item We propose a new paradigm for event extraction -– sequence-to-structure generation, which can directly extract events from the text in an end-to-end manner.
    By uniformly modeling all tasks in a single model and universally predicting different labels, our method is effective, data-efficient, and easy to implement.
    \item We design an effective sequence-to-structure architecture, which is enhanced with a constrained decoding algorithm for event knowledge injection during inference and a curriculum learning algorithm for efficient model learning.
    \item Many information extraction tasks can be formulated as structure prediction tasks. Our sequence-to-structure method can motivate the learning of other information extraction models.
\end{enumerate}

\section{\modelname: End-to-end Event Extraction as Controllable Generation} \label{sec:model}

Given the token sequence $x=x_{1}, ..., x_{|x|}$ of the input text, \modelname\, directly generate the event structures $E=e_{1},...,e_{|E|}$ via an encoder-decoder architecture.
For example, in \figurename~\ref{fig:motivation},  \modelname\, take the raw text as input and output two event records including \{Type: \eventtpye{Transport}, Trigger: returned, Arg1 Role: \eventtpye{Artifact}, Arg1: The man, ...\} and \{Type: \eventtpye{Arrest-Jail}, Trigger: capture, ..., Arg2 Role: \eventtpye{Agent}, Arg2: bounty hunters, ...\}.

For end-to-end event extraction, \modelname\, first encodes input text, then generates the linearized structure using the constrained decoding algorithm.
In the following, we first introduce how to reformulate event extraction as structure generation via structure linearization, then describe the sequence-to-structure model and the constrained decoding algorithm.

\begin{figure*}[!tphb] 
  \centering
\begin{subfigure}[b]{0.32\textwidth}
    \centering
    \includegraphics[width=\textwidth]{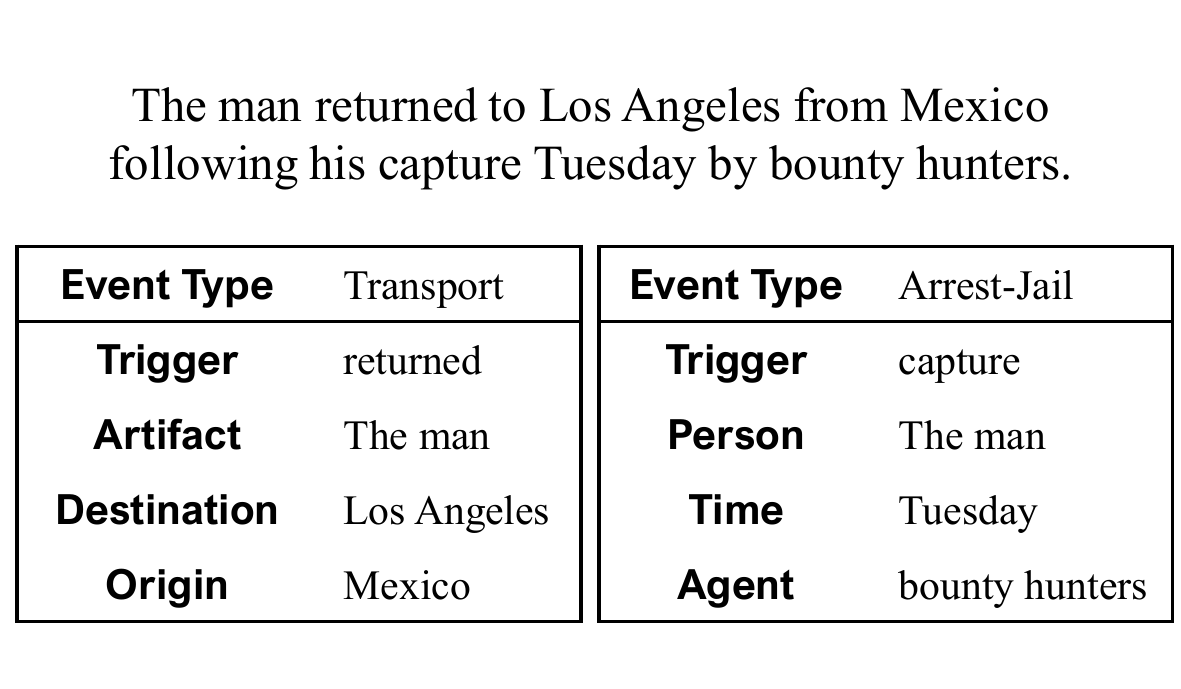}
    \caption{Record format.}
    \label{fig:table_format}
\end{subfigure}
\begin{subfigure}[b]{0.32\textwidth}
    \centering
    \includegraphics[width=\textwidth]{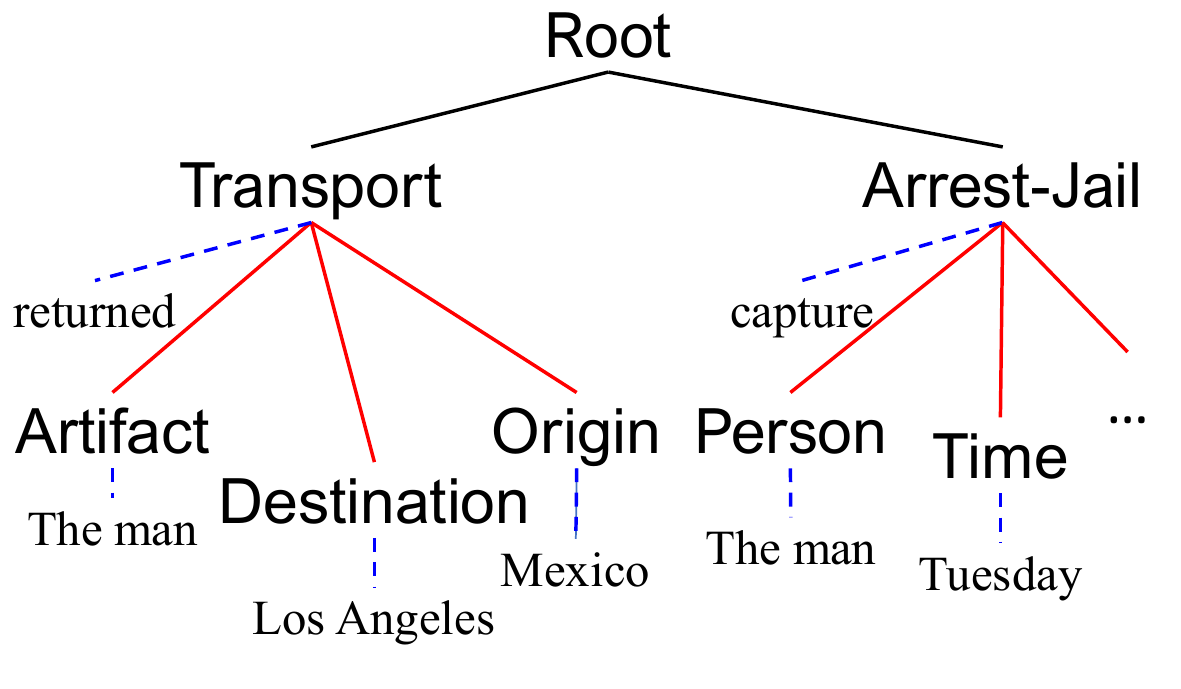}
    \caption{Tree format.}
    \label{fig:event_tree}
\end{subfigure}
\begin{subfigure}[b]{0.32\textwidth}
\begin{minipage}{.30\textwidth}
\begin{Verbatim}[fontsize=\small]
((Transport returned
  (Artifact The man)
  (Destination Los Angeles)
  (Origin Mexico))
 (Arrest-Jail capture
  (Person The man)
  (Time Tuesday)
  (Agent bounty hunters))
\end{Verbatim}
\end{minipage}
\caption{Linearized format.}
\label{fig:linearized_form}
\end{subfigure}

\caption{
  Examples of three event representations.
  The {\color{red} red solid line} indicates the event-role relation; the {\color{blue} blue dotted line} indicates the label-span relation where the head is a label and the tail is a text span.
  For example, ``\eventtpye{Transport}-returned'' is a label-span relation edge, which head is ``\eventtpye{Transport}'' and tail is ``returned''.
}
\label{fig:event_structured_representation}
\end{figure*}

\subsection{Event Extraction as Structure Generation} \label{sec:structure_generation}

This section describes how to linearize event structure so that events can be generated in an end-to-end manner.
Specifically, the linearized event representations should:
(1) be able to express multiple event records in a text as one expression;
(2) be easy to reversibly converted to event records in a deterministic way;
(3) be similar to the token sequence of general text generation tasks so that text generation models can be leveraged and transferred easily.

Concretely, the process of converting from record format to linearized format is shown in \figurename~\ref{fig:event_structured_representation}.
We first convert event records (\figurename~\ref{fig:table_format}) into a labeled tree (\figurename~\ref{fig:event_tree}) by:
1) first labeling the root of the tree with the type of event (Root - Transport, Root - Arrest-Jail),
2) then connecting multiple event argument role types with event types (Transport - Artifact, Transport - Origin, etc.),
and 3) finally linking the text spans from the raw text to the corresponding nodes as leaves (Transport - returned, Transport - Origin - Mexico, Transport - Artifact - The man, etc.).
Given the converted event tree, we linearize it into a token sequence (\figurename~\ref{fig:linearized_form}) via depth-first traversal \citep{NIPS2015_277281aa}, where ``('' and ``)'' are structure indicators used to represent the semantic structure of linear expressions.
The traversal order of the same depth is the order in which the text spans appear in the text, e.g., first ``return'' then ``capture'' in \figurename~\ref{fig:event_tree}.
Noted that each linearized form has a virtual root -- \textit{Root}.
For a sentence that contains multiple event records, each event links to \textit{Root} directly.
For a sentence that doesn't express any event, its tree format will be linearized as ``()''.

\subsection{Sequence-to-Structure Network}\label{sec:seq-to-structure}

Based on the above linearization strategy, \modelname\, generates the event structure via a transformer-based encoder-decoder architecture \citep{NIPS2017_3f5ee243}.
Given the token sequence $x=x_{1}, ..., x_{|x|}$ as input, \modelname\, outputs the linearized event representation $y=y_{1}, ..., y_{|y|}$.
To this end, \modelname\, first computes the hidden vector representation $\mathbf{H}=\mathbf{h}_{1},...,\mathbf{h}_{|x|}$ of the input via a multi-layer transformer encoder: 
\begin{equation} \label{equ:encoder}
    \mathbf{H} = \text{Encoder} (x_{1}, ..., x_{|x|})
\end{equation}
where each layer of $\text{Encoder}(\cdot)$ is a transformer block with the multi-head attention mechanism.

After the input token sequence is encoded, the decoder predicts the output structure token-by-token with the sequential input tokens' hidden vectors.
At the step $i$ of generation, the self-attention decoder predicts the $i$-th token $y_{i}$ in the linearized form and decoder state $\mathbf{h}^{d}_{i}$ as:
\begin{equation} \label{equ:decoder}
    y_{i}, \mathbf{h}^{d}_{i} = \text{Decoder}([\mathbf{H}; \mathbf{h}^{d}_{1},...,\mathbf{h}^{d}_{i-1}], y_{i-1})
\end{equation}
where each layer of $\text{Decoder}(\cdot)$ is a transformer block that contains self-attention with decoder state $\mathbf{h}^{d}_{i}$ and cross-attention with encoder state $\mathbf{H}$.

The generated output structured sequence starts from the start token ``$\langle$bos$\rangle$'' and ends with the end token ``$\langle$eos$\rangle$''. The conditional probability of the whole output sequence $p(y|x)$ is progressively combined by the probability of each step $p(y_{i}|y_{<i},x)$:
\begin{equation}
    \label{equ:condition_prob}
    p(y|x) = \prod_{i}^{|y|} p(y_{i}|y_{<i},x)
\end{equation}
where $y_{<i}=y_{1} ... y_{i-1}$, and $p(y_{i}|y_{<i},x)$ is the probability over the target vocabulary $\mathcal{V}$ normalized by softmax($\cdot$) .

Because all tokens in linearized event representations are also natural language words, we adopt the pre-trained language model T5 \cite{Raffel:JMLR2020:T5} as our transformer-based encoder-decoder architecture.
In this way, the general text generation knowledge can be directly reused.

\subsection{Constrained Decoding} \label{sec:constrained_decoding}

Given the hidden sequence $\mathbf{H}$, the sequence-to-structure network needs to generate the linearized event representations token-by-token.
One straightforward solution is to use a greedy decoding algorithm, which selects the token with the highest predicted probability $p(y_{i}|y_{<i},x)$ at each decoding step $i$.
Unfortunately, this greedy decoding algorithm cannot guarantee the generation of valid event structures.
In other words, it could end up with invalid event types, mismatch of argument-type, and incomplete structure.
Furthermore, the greedy decoding algorithm ignores the useful event schema knowledge, which can be used to guide the decoding effectively.
For example, we can constrain the model to only generate event type tokens in the type position.

To exploit the event schema knowledge, we propose to employ a trie-based constrained decoding algorithm \citep{chen-etal-2020-parallel,decao2020autoregressive} for event generation.
During constrained decoding, the event schema knowledge is injected as the prompt of the decoder and ensures the generation of valid event structures.

\begin{figure}[!tphb] 
\centering

\begin{subfigure}[b]{0.49\textwidth}
\resizebox{\textwidth}{!}{
\begin{tikzpicture}[grow=right, node distance=0.2cm]
    \tikzstyle{level 1}=[level distance=1cm]
    \tikzstyle{level 5}=[level distance=0.9cm]
    \tikzstyle{token} = [text centered]
    \node[token] {$\langle$bos$\rangle$}
    child {
     node[token] {(}
        child {
            node[token] {)}
                child { node[token] {$\langle$eos$\rangle$} edge from parent }
            edge from parent 
        }
        child {
            node[token] {(}
            child {
                    node[token] {$\red{\mathcal{T}}$}
                    child {
                        node[token] {${\color{blue}\mathcal{S}}$}
                        child { 
                            node[token] {)} 
                            child { node[token] {)} 
                                child { node[token] {$\langle$eos$\rangle$} edge from parent }
                                edge from parent 
                                }
                            edge from parent 
                        }
                        child { 
                            node[token] {(}
                            child {
                                node[token] {${\color{red}\mathcal{R}}$}
                                child {
                                    node[token] {${\color{blue}\mathcal{S}}$}
                                    child {
                                        node[token] {)}
                                            child { 
                                                node[token] {(} 
                                                child {node[token] {...} edge from parent}
                                                edge from parent 
                                            }
                                            child { 
                                                node[token] {)} 
                                                child {node[token] {...} edge from parent}
                                                edge from parent 
                                            }
                                        edge from parent
                                    }
                                    edge from parent 
                                }
                                edge from parent
                            }
                            edge from parent
                        }
                        edge from parent
                    }
                    edge from parent 
                }
            edge from parent
        }
    };
\end{tikzpicture}
}
\caption{The trie of event structure.}
\label{fig:trie_integrated}
\end{subfigure}

\bigskip

\begin{subfigure}[b]{0.35\textwidth}
    \resizebox{\textwidth}{!}{
        \includegraphics[width=\textwidth]{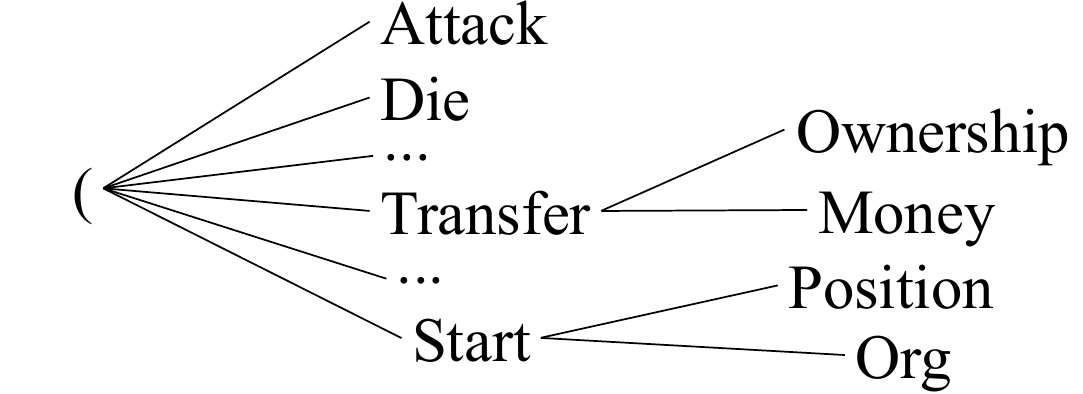}
    }
    \caption{The trie of event type ${\color{red}\mathcal{T}}$.}
    \label{fig:event_type_decoding}
\end{subfigure}

\caption{
  The prefix tree (trie) of the constrained decoding algorithm for controllable structure generation.
  ${\color{red}\mathcal{T}}$ and ${\color{red}\mathcal{R}}$ indicate the label name of event type and argument role.
  ${\color{blue}\mathcal{S}}$ indicates the text span in the raw text, which is the event trigger or argument mention of the extracted event.
}
\label{fig:decoding_stage}
\end{figure}
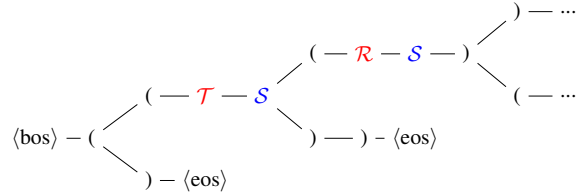
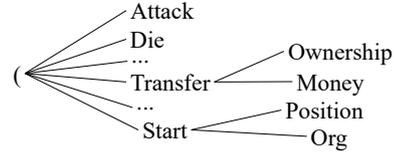

Concretely, unlike the greedy decoding algorithm that selects the token from the whole target vocabulary $\mathcal{V}$ at each step, our trie-based constrained decoding method dynamically chooses and prunes a candidate vocabulary $\mathcal{V'}$ based on the current generated state.
A complete linearized form decoding process can be represented by executing a trie tree search, as shown in \figurename~\ref{fig:trie_integrated}.
Specifically, each generation step of \modelname\, has three kinds of candidate vocabulary $\mathcal{V'}$:
\begin{itemize}
    \item Event schema: label names of event types ${\color{red}\mathcal{T}}$ and argument roles ${\color{red}\mathcal{R}}$;
    \item Mention strings: event trigger word and argument mention ${\color{blue}\mathcal{S}}$, which is the text span in the raw input;
    \item Structure indicator: ``('' and ``)'' which are used to combine event schemas and mention strings.
\end{itemize}

The decoding starts from the root ``$\langle$bos$\rangle$'' and ends at the terminator ``$\langle$eos$\rangle$''.
At the generation step $i$, the candidate vocabulary $\mathcal{V}'$ is the children nodes of the last generated node.
For instance, at the generation step with the generated string ``$\langle$bos$\rangle$ ('', the candidate vocabulary $\mathcal{V}'$ is \{``('', ``)''\} in \figurename~\ref{fig:trie_integrated}.
When generating the event type name ${\color{red}\mathcal{T}}$, argument role name ${\color{red}\mathcal{R}}$ and text span ${\color{blue}\mathcal{S}}$, the decoding process can be considered as executing search on a subtree of the trie tree.
For example, in \figurename~\ref{fig:event_type_decoding}, the candidate vocabulary $\mathcal{V}'$ for ``( Transfer'' is \{``Ownership'', ``Money''\}.

Finally,  the decoder's output will be transformed to event records and used as final extraction results.

\section{Learning} \label{sec:learning}
This section describes how to learn the \modelname\, neural network in an end-to-end manner.
Our method can be learned using only the coarse parallel text-record annotations, i.e., pairs of $\langle$sentence, event records$\rangle$, with no need for fine-grained token-level annotation used in traditional methods.
Given a training dataset $\mathcal{D} = \{(x_{1}, y_{1}), ...(x_{|\mathcal{D}|}, y_{|\mathcal{D}|})\}$ where each instance is a $\langle$sentence, event records$\rangle$ pair, the learning objective is the negative log-likelihood function as:
\begin{equation}
    \label{equ:base_loss_function}
    \mathcal{L} = - \sum_{(x,y)\in\mathcal{D}} \log p(y|x,\mathbf{\theta})
\end{equation}
where $\mathbf{\theta}$ is model parameters.

Unfortunately, unlike general text-to-text generation models, the learning of sequence-to-structure generation models is more challenging:
1) There is an output gap between the event generation model and the text-to-text generation model.
Compared with natural word sequences, the linearized event structure contains many non-semantic indicators such as ``('' and ``)'', and they don't follow the syntax constraints of natural language sentences.
2) The non-semantic indicators ``('' and ``)'' appear very frequently but contain little semantic information, which will mislead the learning process. 

To address the above challenges, we employ a curriculum learning \citep{icml2009_006,xu-etal-2020-curriculum} strategy.
Specifically, we first train PLMs using simple event substructure generation tasks so that they would not overfit in non-semantic indicators;
then we train the model on the full event structure generation task.

\paragraph{Substructure Learning.}
Because event representations often have complex structures and their token sequences are different from natural language word sequences, it is challenging to train them with the full sequence generation task directly.
Therefore, we first train \modelname\, on simple event substructures.

Specifically, we learn our model by starting from generating only ``(label, span)'' substructures, including ``(type, trigger words)'' and ``(role, argument words)'' substructures.
For example, we will extract substructure tasks in \figurename~\ref{fig:linearized_form} in this stage as: \texttt{(Transport returned) (Artifact The man) (Arrest-Jail capture)}, etc.
We construct a $\langle$sentence, substructures$\rangle$ pair for each extracted substructures, then train our model using the loss in equation \ref{equ:base_loss_function}.

\paragraph{Full Structure Learning.}
After the substructure learning stage, we further train our model for the full structure generation task using the loss in equation \ref{equ:base_loss_function}.
We found the curriculum learning strategy uses data annotation more efficiently and makes the learning process more smooth.

\section{Experiments} \label{sec:experiment}

This section evaluates the proposed \modelname\, model by conducting experiments in both supervised learning and transfer learning settings.

\subsection{Experimental Settings} \label{sec:exp_settings}

\paragraph{Datasets.}
We conducted experiments on the event extraction benchmark -- ACE2005 \citep{ace2005-annotation}, which has 599 English annotated documents and 33 event types.
We used the same split and preprocessing step as the previous work \citep{Zhang:2019:GAIL,wadden-etal-2019-entity,du-cardie-2020-event}, and we denote it as ACE05-EN.

In addition to ACE05-EN, we also conducted experiments on two other benchmarks: ACE05-EN$^+$ and ERE-EN, using the same split and preprocessing step in the previous work \citep{lin-etal-2020-joint}.
Compared to ACE05-EN, ACE05-EN$^+$ and ERE-EN further consider pronoun roles and multi-token event triggers.
ERE-EN contains 38 event categories and 458 documents.

Statistics of all datasets are shown in \tablename~\ref{tab:dataset}.

For evaluation, we used the same criteria in previous work \citep{Zhang:2019:GAIL,wadden-etal-2019-entity,lin-etal-2020-joint}.
Since \modelname\, is a text generation model, we reconstructed the offset of predicted trigger mentions by finding the matched utterance in the input sequence one by one.
For argument mentions, we found the nearest matched utterance to the predicted trigger mention as the predicted offset.

\begin{table}[!t]
    \centering
    \setlength{\belowcaptionskip}{-0.5cm}
    \resizebox{0.48\textwidth}{!}{
        \begin{tabular}{ccccc}
            \toprule
            \textbf{Dataset}                   & \textbf{Split} & \textbf{\#Sents} & \textbf{\#Events} & \textbf{\#Roles} \\
            \midrule
            \multirow{3}[2]{*}{ACE05-EN}       & Train          & 17,172           & 4,202             & 4,859            \\
                                               & Dev            & 923              & 450               & 605              \\
                                               & Test           & 832              & 403               & 576              \\
            \midrule
            \multirow{3}[2]{*}{ACE05-EN$^{+}$} & Train          & 19,216           & 4,419             & 6,607            \\
                                               & Dev            & 901              & 468               & 759              \\
                                               & Test           & 676              & 424               & 689              \\
            \midrule
            \multirow{3}[2]{*}{ERE-EN}         & Train          & 14,736           & 6,208             & 8,924            \\
                                               & Dev            & 1,209            & 525               & 730              \\
                                               & Test           & 1,163            & 551               & 822              \\
            \bottomrule
        \end{tabular}
    }
    \caption{Dataset statistics.}
    \label{tab:dataset}
\end{table}%

\paragraph{Baselines.}

Currently, event extraction supervision can be conducted at two different levels:
1) \textit{Token-level annotation}, which labels each token in a sentence with event labels, e.g., ``The/O dismission/B-End-Position of/O ..'';
2) \textit{Parallel text-record annotation}, which only gives $\langle$sentence, event$\rangle$ pairs but without expensive token-level annotations, e.g., $\langle$The dismission of ..., \{Type: End-Position, Trigger: dismission, ...\}$\rangle$.
Furthermore, some previous works also leverage golden entity annotation for model training, which marks all entity mentions with their golden types, to facilitate event extraction.
Introducing more supervision knowledge will benefit the event extraction but is more label-intensive.
The proposed Text2Event only uses parallel text-record annotation, which makes it more practical in a real-world application.

\begin{table*}[!t]
  \centering
  \begin{tabular}{r|ccc|ccc|c}
    \toprule
    \multicolumn{1}{c}{\multirow{2}[0]{*}{\textbf{Models}}} & \multicolumn{3}{c}{\textbf{Trig-C}} & \multicolumn{3}{c}{\textbf{Arg-C}} & \multirow{2}[0]{*}{\textbf{PLM}}                                                               \\
    \cmidrule(r){2-4} \cmidrule(r){5-7}
                                                            & \textbf{P}                          & \textbf{R}                         & \textbf{F1}                      & \textbf{P} & \textbf{R} & \textbf{F1} &                     \\
    \midrule
    \multicolumn{8}{c}{Models using Token Annotation + Entity Annotation}                                                                                                                                                               \\
    \midrule
    Joint3EE \citep{Nguyen-Nguyen:2019:AAAI2019}            & 68.0                                & 71.8                               & 69.8                             & 52.1       & 52.1       & 52.1        & -                   \\
    DYGIE++  \citep{wadden-etal-2019-entity}                & -                                   & -                                  & 69.7                             & -          & -          & 48.8        & BERT-large          \\
    GAIL \citep{Zhang:2019:GAIL}                            & 74.8                                & 69.4                               & 72.0                             & 61.6       & 45.7       & 52.4        & ELMo                \\
    OneIE$_{\text{w/o Global}}$ \citep{lin-etal-2020-joint} & -                                   & -                                  & 73.5                             & -          & -          & 53.9        & BERT-large          \\
    OneIE \citep{lin-etal-2020-joint}                       & -                                   & -                                  & 74.7                             & -          & -          & 56.8        & BERT-large          \\
    \midrule
    \multicolumn{8}{c}{Models using Token Annotation}                                                                                                                                                                                   \\
    \midrule
    EEQA \citep{du-cardie-2020-event}                       & 71.1                                & 73.7                               & 72.4                             & 56.8       & 50.2       & 53.3        & 2$\times$BERT-base  \\
    MQAEE  \citep{li-etal-2020-event}                       & -                                   & -                                  & 71.7                             & -          & -          & 53.4        & 3$\times$BERT-large \\
    \midrule
    \multicolumn{8}{c}{Generation-based Baselines using Token Annotation}                                                                                                                                                               \\
    \midrule
    TANL \citep{paolini2021structured}                      & -                                   & -                                  & 68.4                             & -          & -          & 47.6        & T5-base             \\
    Multi-Task TANL \citep{paolini2021structured}           & -                                   & -                                  & 68.5                             & -          & -          & 48.5        & T5-base             \\
    \midrule
    \multicolumn{8}{c}{Our Model using Parallel Text-Record Annotation}                                                                                                                                                                 \\
    \midrule
    \modelname\,                                            & 67.5                                & 71.2                               & 69.2                             & 46.7       & 53.4       & 49.8        & T5-base             \\
    \modelname\,                                            & 69.6                                & 74.4                               & 71.9                             & 52.5       & 55.2       & 53.8        & T5-large            \\
    \bottomrule
  \end{tabular}
  \caption{Experiment results on ACE05-EN.
    Trig-C indicates trigger identification and classification.
    Arg-C indicates argument identification and classification.
    PLM represents the pre-trained language models used by each model.
  }
  \label{tab:overall_ace2005}%
\end{table*}%

To verify \modelname, we compare our method with the following groups of baselines:

1. Baselines using token annotation:
\textit{TANL} is the SOTA sequence generation-based method that models event extraction as a trigger-argument pipeline manner \citep{paolini2021structured};
\textit{Multi-task TANL} extends \textit{TANL} by transferring structure knowledge from other tasks;
\textit{EEQA} \citep{du-cardie-2020-event} and \textit{MQAEE} \citep{li-etal-2020-event} are QA-based models which use machine reading comprehension model for trigger detection and argument extraction.

2. Baselines using both token annotation and entity annotation:
\textit{Joint3EE} is a joint entity, trigger, argument extraction model based on the shared hidden representations \citep{Nguyen-Nguyen:2019:AAAI2019};
\textit{DYGIE++} is a BERT-based model which captures both within-sentence and cross-sentence context \citep{wadden-etal-2019-entity};
\textit{GAIL} is an inverse reinforcement learning-based joint entity and event extraction model \citep{Zhang:2019:GAIL};
\textit{OneIE} is an end-to-end IE system which employs global feature and beam search to extract globally optimal event structures \citep{lin-etal-2020-joint}.

\paragraph{Implementations.}
We optimized our model using label smoothing \citep{7780677,MullerKH19} and AdamW \citep{loshchilov2018decoupled} with learning rate=5e-5 for T5-large, 1e-4 for T5-base.
For curriculum learning, the epoch of substructure learning is 5, and full structure learning is 30.
We conducted each experiment on a single NVIDIA GeForce RTX 3090 24GB.
Due to GPU memory limitation, we used different batch sizes for different models: 8 for T5-large and 16 for T5-base; and truncated the max length of raw text to 256 and linearized form to 128 during training.
We added the task name as the prefix for the T5 default setup.

\subsection{Results in Supervised Learning Setting} \label{sec:supervised_learning_results}

\tablename~\ref{tab:overall_ace2005} presents the performance of all baselines and \modelname\, on ACE05-EN.
And \tablename~\ref{tab:results_on_new_benchmark} shows the performance of SOTA and \modelname\, on ACE05-EN$^{+}$ and ERE-EN.
We can see that:

1) \textit{By uniformly modeling all tasks in a single model and predicting labels universally, \modelname\, can achieve competitive performance with weaker supervision and simpler architecture.}
Our method, only using the weak parallel text-record annotations, surpasses most of the baselines using token and entity annotations and achieves competitive performance with SOTA.
Furthermore, using the simple encoder-decoder architecture, \modelname\, outperforms most of the counterparts with complicated architectures.

2) \textit{By directly generating event structure from the text, \modelname\, can significantly outperform sequence generation-based methods.}
Our method improves Arg-C F1 by 4.6\% and 2.7\% over the SOTA generation baseline and its extended multi-task TANL.
Compared with sequence generation, structure generation can be effectively guided using event schema knowledge during inference, and there is no need to generate irrelevant information.

3) \textit{By uniformly modeling and sharing information between different tasks and labels, the sequence-to-structure framework can achieve robust performance. }
From \tablename~\ref{tab:overall_ace2005} and \tablename~\ref{tab:results_on_new_benchmark}, we can see that the performance of OneIE decreases on the harder dataset ACE05-EN$^{+}$, which has more pronoun roles and multi-token triggers.
By contrast, the performance of \modelname\, remains nearly the same on ACE05-EN.
We believe this may be because the proposed sequence-to-structure model is a universal model that doesn't specialize in labels and can better share information between different labels.

\begin{table}[!t]
  \centering
    \resizebox{0.49\textwidth}{!}{
    \begin{tabular}{c|ccc|ccc}
      \toprule
      \multicolumn{1}{c}{\multirow{2}[4]{*}{\textbf{Datasets}}}  & \multicolumn{3}{c}{\textbf{Trig-C}} & \multicolumn{3}{c}{\textbf{Arg-C}} \\
      \cmidrule(r){2-4} \cmidrule(r){5-7}
          & \textbf{P} & \textbf{R} & \textbf{F1} & \textbf{P} & \textbf{R} & \textbf{F1} \\
      \midrule
        \multicolumn{7}{c}{SOTA (Token + Entity Annotation)} \\
        \midrule
    ACE05-EN$^{+}$ & - & - & 72.8  & - & - & 54.8  \\
    ERE-EN$^{*}$ & 56.9  & 58.7  & 57.8  & 51.9  & 47.8  & 49.8  \\
    \midrule
          \multicolumn{7}{c}{\modelname\, (Parallel Text-Record Annotation)} \\
          \midrule
          ACE05-EN$^{+}$ & 71.2 & 72.5 & 71.8 & 54.0 & 54.8 & 54.4  \\
          ERE-EN & 59.2 & 59.6 & 59.4 & 49.4 & 47.2 & 48.3  \\
    \bottomrule
    \end{tabular}%
    }
    \caption{
        Experiment results on ACE05-EN$^{+}$ and ERE-EN.
        SOTA indicates the state-of-the-art system -- OneIE.
        * The result of SOTA for ERE-EN is reproduced by the official release code because of the slightly different dataset statistic result on ERE-EN.
      }
  \label{tab:results_on_new_benchmark}%
\end{table}%

\subsection{Results in Transfer Learning Setting} \label{sec:transfer_learning_results}
\modelname\, is a universal model, therefore can facilitate the knowledge transfer between different labels.
To verify the transfer ability of \modelname, we conducted experiments in the transfer learning setting, and the results are shown in \tablename~\ref{tab:transfer-learning}.
Specifically, we first randomly split the sentences which length larger than 8 in ACE05-EN$^{+}$ into two equal-sized subsets \textit{src} and \textit{tgt}:
\textit{src} only retains the annotations of the top 10 frequent event types,
and \textit{tgt} only retains the annotations of the remaining 23 event types.
For both \textit{src} and \textit{tgt}, we use 80\% of the dataset for model training and 20\% for evaluation.
For transfer learning,  We first pre-trained an event extraction model on the \textit{src} dataset, then fine-tuned the pre-trained model for extracting the new event types in \textit{tgt}.
From \tablename~\ref{tab:transfer-learning}, we can see that:

\begin{table}[!t]
    \centering
    \resizebox{0.48\textwidth}{!}{
    \begin{tabular}{r|ccc|ccc}
      \toprule
      \multicolumn{1}{c}{\multirow{2}[0]{*}{ \textbf{Settings} }}  & \multicolumn{3}{c}{\textbf{Trig-C}} & \multicolumn{3}{c}{\textbf{Arg-C}} \\
      \cmidrule(r){2-4} \cmidrule(r){5-7} 
      & \textbf{P} & \textbf{R} & \textbf{F1} & \textbf{P} & \textbf{R} & \textbf{F1} \\
      \midrule
      \multicolumn{7}{c}{OneIE (Token + Entity Annotation)} \\
      \midrule
      Non-transfer  & 78.1 & 62.3 & 69.3 & 50.9 & 37.9 & 43.5 \\
      Transfer   & 78.9 & 61.7 & 69.2 & 57.1 & 40.0 & 47.0 \\
      \textit{Gain} & & & -0.1 & & & +3.5 \\
      \midrule
      \multicolumn{7}{c}{EEQA (Token Annotation)} \\
      \midrule
      Non-transfer  & 69.9 & 67.3 & 68.6 & 36.5 & 37.4 & 36.9 \\
      Transfer    & 79.5 & 61.7 & 69.5 & 33.9 & 41.2 & 37.2 \\
      \textit{Gain} & & & +0.9 & & & +0.3 \\
      \midrule
      \multicolumn{7}{c}{\modelname\, (Parallel Text-Record Annotation)} \\
      \midrule
      Non-transfer  & 79.4 & 61.1 & 69.0 & 58.4 & 40.9 & 48.0  \\
      Transfer    & 82.1 & 65.3 & 72.7 & 58.8 & 45.4 & 51.2  \\
      \textit{Gain}  & & & +3.7 & & & +3.2 \\
      \bottomrule
    \end{tabular}%
    }
    \caption{
      Experiment results on the $tgt$ subset of ACE05-EN$^{+}$ in the transfer learning setting.
      }
    \label{tab:transfer-learning}%
\end{table}%

1) \textit{Data-efficient \modelname\, can make better use of supervision signals.}
Even training on $tgt$ from scratch, the proposed method also outperforms strong baselines.
We believe that this may because baselines using token and entity annotation require massive fine-grained data for model learning.
Different from baselines, \modelname\, uniformly models all subtasks, thus the knowledge can be seamlessly transferred, which is more data-efficient.

2) \textit{\modelname\, can effectively transfer knowledge between different labels.}
Compared with the non-transfer setting, which is directly trained on $tgt$ training set, the transfer setting of \modelname\, can achieve significant F1 improvements of 3.7 and 3.2 on Trig-C and Arg-C, respectively.
By contrast, the other two baselines cannot obtain significant F1 improvements of both Trig-C and Arg-C via transfer learning.
Note that the information of entity annotation is shared across \textit{src} and \textit{tgt}.
As a result, OneIE can leverage such information to better argument prediction even with worse trigger prediction.
However, even without using entity annotation, the proposed method can still achieve a similar improvement in the transfer learning setting.
This is because the labels are provided universally in \modelname, so the parameters are not label-specific.

\subsection{Detailed Analysis}

This section analyzes the effects of event schema knowledge, constrained decoding, and curriculum learning algorithm in \modelname.
We designed four ablated variants based on T5-base:
\begin{itemize}
    \item ``\modelname'' is the base model that is directly trained with the full structure learning.
    \item ``+ CL'' indicates training \modelname\, with the proposed curriculum learning algorithm.
    \item ``w/o CD'' discards the constrained decoding during inference and generates event structures as an unconstrained generation model.
    \item ``w/o ES'' replaces the names of event types and roles with meaningless symbols, which is used to verify the effect of event schema knowledge.
\end{itemize}
\begin{table}[!t]
  \centering
  \setlength{\belowcaptionskip}{-0.4cm}
  \resizebox{0.49\textwidth}{!}{
    \begin{tabular}{lcccc}
      \toprule
      \multicolumn{1}{c}{\textbf{Trig-C F1}} & 1\%  & 5\%  & 25\% & 100\%                \\
      \midrule
      \modelname\, + CL                       & 24.6 & 52.8 & 65.5 & 71.4                 \\
      \modelname\,                           & 17.9 & 52.1 & 65.0 & 69.6                 \\
      \quad w/o CD                           & 13.2 & 46.8 & 64.3 & 68.6                 \\
      \quad w/o ES                           & 0.0  & 24.3 & 31.6 & 55.5                 \\
      \midrule \midrule
      \multicolumn{1}{c}{\textbf{Arg-C F1}}  & 1\%  & 5\% & 25\% & 100\% \\
      \midrule
      \modelname\, + CL                      & 8.6  & 33.6 & 44.0 & 53.3 \\ 
      \modelname\,                           & 3.7  & 30.9 & 44.7 & 52.6\\ 
      \quad w/o CD                           & 2.3  & 27.3 & 44.4 & 52.3\\ 
      \quad w/o ES                           & 0.0  & 7.0  & 8.2  & 28.9\\ 
      \bottomrule
    \end{tabular}
  }
  \caption{
    Experiment results of variants trained with different-sized training set on the development set of ACE05-EN.
  }
  \label{tab:ablation_analysis}%
\end{table}

\tablename~\ref{tab:ablation_analysis} shows the results on the development set of ACE05-EN using different training data sizes.
We can see that:
1) \textit{Constrained decoding can effectively guide the generation with event schemas, especially in low-resource settings.}
Comparing to ``w/o CD'', constrained decoding improves the performance of \modelname, especially in low-resource scenarios, e.g., using 1\%, 5\% training set.
2) \textit{Curriculum learning is useful for model learning.}
Substructure learning improves 4.7\% Trig-C F1 and 5.8\% Arg-C F1 on average.
3) \textit{It is crucial to encode and generate event labels as words, rather than meaningless symbols.}
Because by encoding labels as natural language words, our method can effectively transfer knowledge from pre-trained language models.

\section{Related Work} \label{sec:relatedwork}
Our work is a synthesis of two research directions: event extraction and structure prediction via neural generation model.

Event extraction has received widespread attention in recent years, and mainstream methods usually use different strategies to obtain a complete event structure.
These methods can be divided into: 
1) pipeline classification \citep{ahn:2006:ARTE,Ji-and-Grishman:2008:ACL2008,liao-grishman-2010-using,hong-etal-2011-using,hong-etal-2018-self,huang-riloff:2012:aaai,chen-etal-2015-event,sha-etal-2016-rbpb,lin-etal-2018-nugget,yang-etal-2019-exploring-pre,wang-etal-2019-hmeae,ma-etal-2020-resource,zhang-etal-2020-two},
2) multi-task joint models \citep{mcclosky-etal-2011-event,li-etal-2013-joint,li-etal-2014-constructing,yang-mitchell:2016:N16-1,nguyen-cho-grishman:2016:N16-1,liu-etal-2018-jointly,zhang-etal:2019:IJCAI,zheng-etal-2019-doc2edag}, 
3) semantic structure grounding \citep{huang-etal-2016-liberal,huang-etal-2018-zero,zhang2020unsupervised},
and 4) question-answering \citep{chen-etal-2020-reading,du-cardie-2020-event,li-etal-2020-event,liu-etal-2020-event}.

Compared with previous methods, we model all subtasks of event extraction in a uniform sequence-to-structure framework, which leads to better decision interactions and information sharing.
The neural encoder-decoder generation architecture \citep{NIPS2014_a14ac55a,bahdanau2014neural} has shown its strong structure prediction ability and has been widely used in many NLP tasks, such as 
machine translation \citep{kalchbrenner-blunsom-2013-recurrent-continuous},
semantic parsing \citep{dong-lapata-2016-language,song-etal-2020-structural},
entity extraction \citep{strakova-etal-2019-neural},
relation extraction \citep{zeng-etal-2018-extracting,zhang-etal-2020-minimize},
and aspect term extraction \citep{ma-etal-2019-exploring}.
Like \modelname\, in this paper, TANL \citep{paolini2021structured} and GRIT \citep{du2020documentlevel} also employ  neural generation models for event extraction, but they focus on sequence generation, rather than structure generation.
Different from previous works that extract text span via labeling \cite{strakova-etal-2019-neural} or copy/pointer mechanism \cite{zeng-etal-2018-extracting,du2020documentlevel}, \modelname\, directly generate event schemas and text spans to form event records via constrained decoding \citep{decao2020autoregressive,chen-etal-2020-parallel}, which allows \modelname\, to handle various event types and transfer to new types easily.

\section{Conclusions} \label{sec:conclusion}

In this paper, we propose \modelname, a sequence-to-structure generation paradigm for event extraction.
\modelname\, directly learns from parallel text-record annotation and uniformly models all subtasks of event extraction in a sequence-to-structure framework.
Concretely, we propose an effective sequence-to-structure network for event extraction, which is further enhanced by a constrained decoding algorithm for event knowledge injection during inference and a curriculum learning algorithm for efficient model learning.
Experimental results in supervised learning and transfer learning settings show that \modelname\, can achieve competitive performance with the previous SOTA using only coarse text-record annotation.

For future work, we plan to adapt our method to other information extraction tasks, such as \textit{N}-ary relation extraction.

\section*{Acknowledgments}
We sincerely thank the reviewers for their insightful comments and valuable suggestions.
This work is supported by the National Natural Science Foundation of China under Grants no. U1936207 and 61772505,  Beijing Academy of Artiﬁcial Intelligence (BAAI2019QN0502), and in part by the Youth Innovation Promotion Association CAS(2018141).

\bibliography{text2event}
\bibliographystyle{acl_natbib}

\end{document}